\pdfoutput=1

\documentclass[11pt]{article}

\usepackage[]{EMNLP2023}

\usepackage{latexsym}
\usepackage{wrapfig}
\usepackage{bm}
\usepackage{times} 
\usepackage{graphicx}
\usepackage{algorithm}
\usepackage{algpseudocode}
\usepackage{dblfloatfix}
\usepackage{amsmath}
\usepackage{amsfonts}

\usepackage[T1]{fontenc}
\usepackage{color}
\usepackage{hyperref}
\usepackage{pifont}
\usepackage{subcaption}
\usepackage{stackengine}
\usepackage[export]{adjustbox}
\def\delequal{\mathrel{\ensurestackMath{\stackon[1pt]{=}{\scriptstyle\Delta}}}}
\usepackage{tabularray}

\usepackage{enumitem}

\usepackage[T1]{fontenc}

\usepackage[utf8]{inputenc}

\usepackage{microtype}

\usepackage{inconsolata}

\title{
JaSPICE: Automatic Evaluation Metric Using \\ Predicate-Argument Structures for Image Captioning Models
}

\author{Yuiga Wada \\
  Keio University \\
  \texttt{yuiga@keio.jp} \\\And
  Kanta Kaneda \\
  Keio University \\
  \texttt{k.kaneda@keio.jp} \\\And
  Komei Sugiura \\
  Keio University \\
  \texttt{komei.sugiura@keio.jp} \\}

\begin{document}

\maketitle
\thispagestyle{empty}
\pagestyle{empty}

\begin{abstract}
Image captioning studies heavily rely on automatic evaluation metrics such as BLEU and METEOR.
However, such $n$-gram-based metrics have been shown to correlate poorly with human evaluation, leading to the proposal of alternative metrics such as SPICE for English; however, no equivalent metrics have been established for other languages.
Therefore, in this study, we propose an automatic evaluation metric called JaSPICE, which evaluates Japanese captions based on scene graphs.
The proposed method generates a scene graph from dependencies and the predicate-argument structure, and extends the graph using synonyms.
We conducted experiments employing 10 image captioning models trained on STAIR Captions and PFN-PIC 
and constructed the \textit{Shichimi} dataset, which contains 103,170 human evaluations.
The results showed that our metric outperformed the baseline metrics for the correlation coefficient with the human evaluation.
\end{abstract}

\vspace{-2mm}
\section{Introduction
\label{intro}
}
\vspace{-1mm}

Image captioning has been extensively studied and applied to various applications in society, such as generating fetching instructions for robots, assisting blind people, and answering questions from images\cite{mabn,alleviating,crt,blind,open-domain-cqg,capwap}.
In this field, it is important that the quality of the generated captions is evaluated appropriately.
However, researchers have reported that automatic evaluation metrics based on $n$-grams do not correlate well with human evaluation\cite{spice}.
Alternative metrics that do not rely on $n$-grams have been proposed for English (e.g., SPICE\cite{spice}); however, they are not fully applicable to all languages.
Therefore, developing an automatic evaluation metric that correlates well with human evaluation for image captioning models in languages other than English would be beneficial.

\begin{figure}[t]
    \centering
    \includegraphics[width=\linewidth]{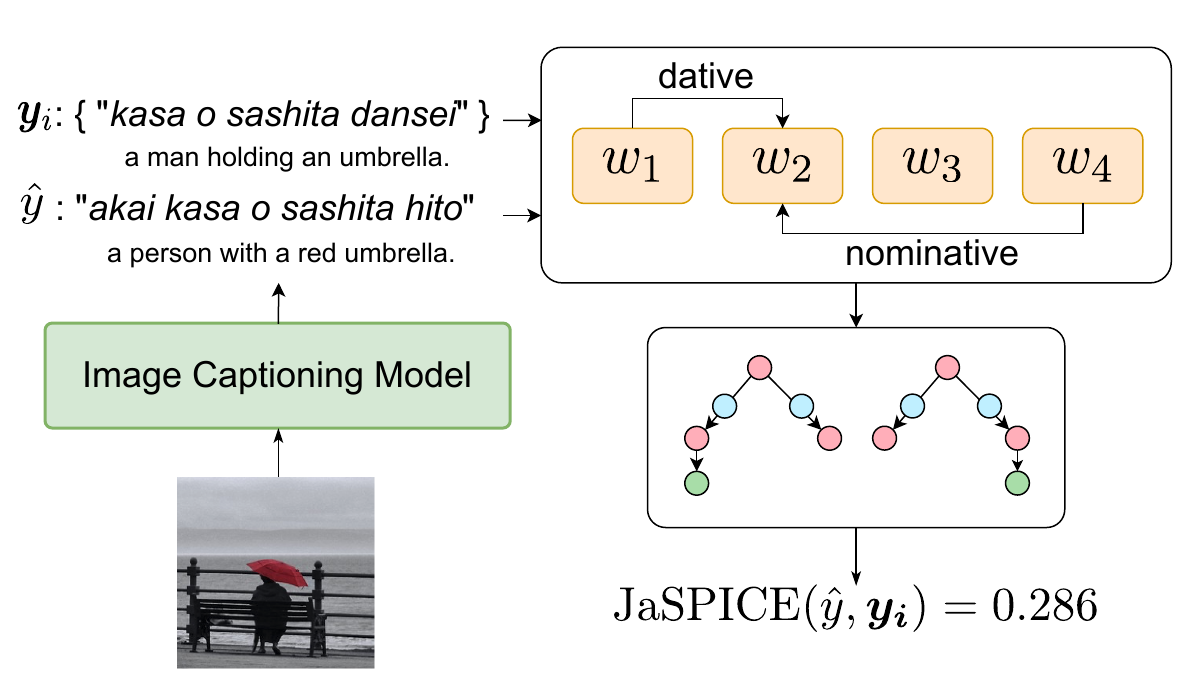}
    \caption{Overview of JaSPICE\footnotemark. Given a candidate caption and reference captions, our method parses the scene graph from the PAS and dependencies, and then computes a score that represents the similarity between the candidate and the references by matching both graphs.
}
    \label{fig:eye_catch}
    \vspace{-0.75cm}
\end{figure}

\footnotetext{Project page: \url{https://yuiga.dev/jaspice/en}}

SPICE is a standard metric for image captioning in English and evaluates captions based on scene graphs.
SPICE uses Universal Dependency (UD) \cite{ud} to generate scene graphs; however, UD can only extract basic dependencies and cannot handle complex relationships. 
In the case of Japanese, the phrase ``A \textit{no} B'' \cite{anob}, which is composed of the nouns A and B has multiple semantic relations, which makes the semantic analysis of such phrases a challenging problem.
For example, in the noun phrase ``\textit{kinpatsu no dansei}'' (``a blond man''), ``blond'' (A) is an attribute of  ``a man'' (B) and ``a man'' (B) is an object, whereas the noun phrase ``\textit{dansei no kuruma}'' (``man's car'') represents the relation of a ``car'' (B) being owned by a ``man'' (A), and dependency parsing using UD cannot accurately extract the relationship between A and B.
Given that UD cannot handle complex relationships and is therefore not suitable for constructing scene graphs, directly applying SPICE to the evaluation of Japanese captions poses challenges.
Furthermore, problem settings exist that are difficult to evaluate using SPICE simply computed from an English translation (e.g., TextCaps\cite{textcaps}).

To address these issues, we propose JaSPICE, which is an automatic evaluation metric for image captioning models in Japanese.
JaSPICE is computed from scene graphs generated from dependencies and the predicate-argument structure (PAS) and can therefore take complex relationships into account.

Fig. \ref{fig:eye_catch} illustrates our JaSPICE approach, where the main idea is that we first parse the scene graph from the PAS and dependencies, and then computes a score that represents the similarity between the candidate caption and the reference captions by matching both graphs.
For example, given the candidate caption ``\textit{akai kasa o sashita hito}'' (``a person with a red umbrella'') and the reference caption ``\textit{kasa o sashita dansei}'' (``a man with an umbrella''), our method parses the scene graph and computes a score by matching both graphs.

Our method differs from existing methods because it generates scene graphs based on dependencies and the PAS and uses synonym sets for the evaluation so that it can evaluate image captioning models in Japanese.
It is expected that appropriate scene graphs can be generated by reflecting dependencies and PAS in scene graphs.
It is also expected that the use of synonym sets will improve the correlation of metrics with human evaluation because it considers the matching of synonyms that do not match on the surface.

The main contributions are as follows:
\begin{itemize}
    \vspace{-2mm}
    \setlength{\parskip}{0cm}
    \setlength{\itemsep}{0cm}
    \item We propose JaSPICE, which is an automatic evaluation metric for image captioning models in Japanese.
    \item Unlike SPICE which uses UD, JaSPICE generates scene graphs based on dependencies and the PAS.
    \item We introduce a graph extension using synonym relationships to take synonyms into account in the evaluation.
    \item We constructed the \textit{Shichimi} dataset, which contains a total of 103,170 human evaluations collected from 500 evaluators.
\end{itemize}

\section{Related Work
\label{related}
}

\subsection{Image Captioning and Its Applications}
Many studies have been conducted in the field of image captioning\cite{sat,ort,m2trm,dlct,ng-etal-2021-understanding,er-san}.
For instance, \cite{ic-survey} is a survey paper that provides a comprehensive overview of image caption generation, including models, standard datasets, and evaluation metrics. Specifically, various automatic evaluation metrics such as embedding-based metrics\cite{wmd} and learning-based metrics\cite{bert-score} have been comprehensively summarized.

Standard datasets for English image captioning tasks include MS COCO\cite{coco}, Flickr30K\cite{flickr30k} and CC3M \cite{cc3m}. Standard datasets for Japanese image captioning tasks include STAIR Captions\cite{stair} and YJ Captions \cite{yj}, which are based on MS COCO images.


\subsection{Automatic Evaluation Metrics}

Standard automatic metrics for image captioning models include BLEU\cite{bleu}, ROUGE\cite{rouge}, METEOR\cite{meteor} and CIDEr\cite{cider}.
Additionally, SPICE\cite{spice} is considered as a standard metric for evaluating image captioning models in English.

BLEU and METEOR were first introduced for machine translation.
BLEU computes precision using $n$-grams up to four in length, while METEOR favors the recall of matching unigrams. Additionally, ROUGE considers the longest subsequence of tokens that appears in both the candidate and reference captions, and CIDEr uses the cosine similarity between the TF-IDF weighted $n$-grams, thereby considering both precision and recall.
Unlike these metrics, which are based on $n$-grams, SPICE evaluates captions using scene graphs. 

Scene graph has been widely applied to vision-related tasks such as image retrieval \cite{image-retrieval-SG,image-retrieval-SG2,stanford-parser}, image generation\cite{image-generate-SG}, VQA \cite{block,VQA-SG,xnms}, and robot planning\cite{robot-planning} because of their powerful representation of semantic features of scenes.
A scene graph was first proposed in \cite{image-retrieval-SG} as a data structure for describing objects instances in a scene and relationships between objects.
In \cite{image-retrieval-SG}, the authors proposed a method for image retrieval using scene graphs; however, a major shortcoming of their method is that the user needs to enter a query in the form of a scene graph. Therefore, in \cite{stanford-parser}, the authors proposed Stanford Scene Graph Parser, which can parse natural language into scene graphs automatically. \cite{stanford-parser} is one of the early methods for the construction and application of scene graphs.

SPICE parses captions into scene graphs using Stanford Scene Graph Parser and then computes the $F_1$ score based on scene graphs.
Our method differs from SPICE in that our method introduces a novel scene graph parser based on the PAS and dependencies, graph extensions using synonym relationships so that it can evaluate Japanese captions.

\section{Problem Statement
\label{sec:problem}
}

In this study, we focus on the automatic evaluation of image captioning models in Japanese.
The terminology used in this study is defined as follows: 

\begin{itemize}
    \vspace{-2mm}
    \setlength{\parskip}{0cm}
    \setlength{\itemsep}{0cm}
    \item \textbf{Predicate-argument structure (PAS)}: a structure representing the relation between predicates and their arguments in a sentence.
    \item \textbf{Scene graph}: a graph that represents semantic relations between objects in an image. The details are explained in Section \ref{subsec:scene-graph}.
    \vspace{-2mm}
\end{itemize}

Given a candidate caption $\hat{y_i}$ and a set of reference captions $\{y_{i,j}\}_{j=1}^N$, automatic image captioning evaluation metrics compute a score that captures the similarity between $\hat{y_i}$ and $\{y_{i,j}\}_{j=1}^N$. Note that $N$ denotes the number of reference captions. 
We evaluate the proposed metric using its correlation coefficient (Pearson/Spearman/Kendall's correlation coefficient) with human evaluation.
This is because automatic evaluation metrics for image captioning models should correlate highly with human evaluation\cite{spice}.

In this study, we assume that we deal with the automatic evaluation of Japanese image captions.
However, some of the discussion in this study can be applied to other languages.

\begin{figure}[t]
    \centering
    \includegraphics[width=\linewidth]{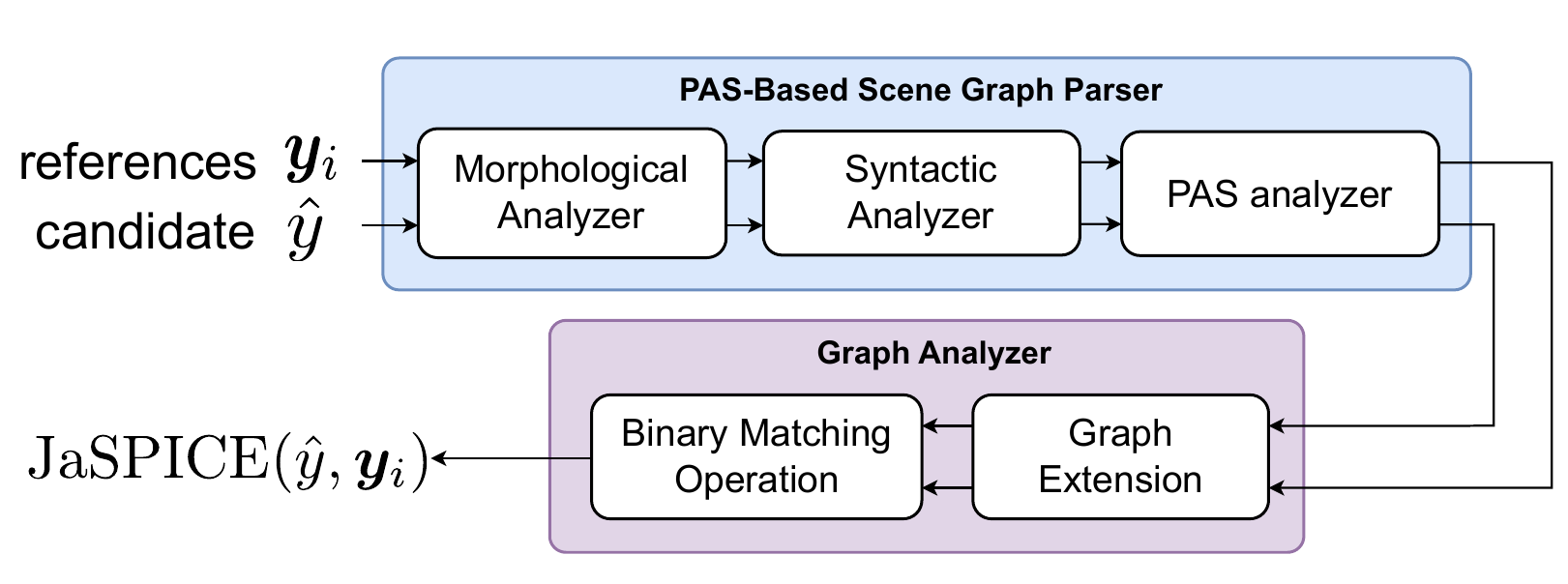}
    \caption{Process diagram of the proposed method. Our method consists of two main modules: PAS-SGP and GA. (i) PAS-SGP generates scene graphs from captions using the PAS and dependencies. (ii) GA performs a graph extension using synonym relationships and then computes the $F_1$ score by matching tuples extracted from the candidate and the reference scene graphs. JaSPICE is easily interpretable because it outputs the score in the range of $[0,1]$.}
    \label{fig:system}
    \vspace{-0.5cm}
\end{figure}

\section{Proposed Method
\label{sec:proposed}
}

For the evaluation of image captioning models, semantic structure is expected to be more effective than $n$-gram because, unlike machine translation, image captioning requires grounding based on the scene and relationships between objects in the image.
Therefore, utilizing the scene graph, which abstracts the lexical and syntactic aspects of natural language, can be beneficial for the evaluation of image captioning models.

In this study, we propose JaSPICE, which is an automatic evaluation metric for image captioning models in Japanese.
JaSPICE is an extension of SPICE\cite{spice} and can evaluate image captioning models in Japanese based on scene graphs.
Although the proposed metric is an extension of SPICE, it also takes into account factors not handled by SPICE, that is, subject completion and the addition of synonymous nodes. Therefore, we believe that the novelty of the proposed metric can be applied to other automatic evaluation metrics.

The main differences between the proposed metric and SPICE are as follows: 
\begin{itemize}
    \vspace{-2mm}
    \setlength{\parskip}{0cm}
    \setlength{\itemsep}{0cm}
    \item Unlike SPICE, JaSPICE generates a scene graph based on dependencies and the PAS.
    \item JaSPICE performs heuristic zero anaphora resolution and graph extension using synonyms.
\end{itemize}

Fig. \ref{fig:system} shows the process diagram of our method.
The proposed method consists of two main modules: PAS-Based Scene Graph Parser (PAS-SGP)  and Graph Analyzer (GA).

\subsection{Scene Graph}
\label{subsec:scene-graph}

The scene graph for a caption $y$ is represented by 
\begin{align}
G(y) = \mathcal{G}  \left\langle O(y),E(y),K(y) \right\rangle , \nonumber
\end{align}
where $O(y)$, $E(y)$, and $K(y)$ denote the set of objects in $y$, the set of relations between objects, and the set of objects with attributes, respectively.
Given that $C$, $R$, and $A$ denote the whole sets of objects, relations, and attributes, respectively, then we can write $O(y) \subseteq C, E(y) \subseteq O(y) \times R \times O(y), K(y) \subseteq  O(y) \times A.$

\begin{figure}[ht]
  \begin{minipage}[b]{0.3\linewidth}
    \centering
    \includegraphics[height=5cm]{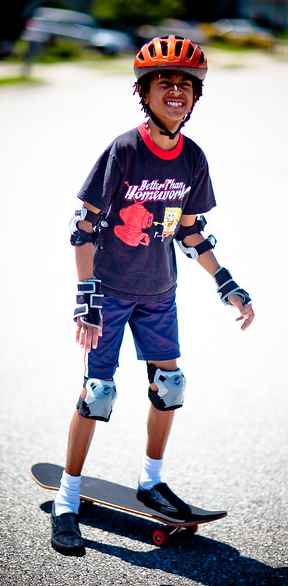}
    \subcaption{}
  \end{minipage}
  \begin{minipage}[b]{0.7\linewidth}
    \centering
    \includegraphics[height=5cm]{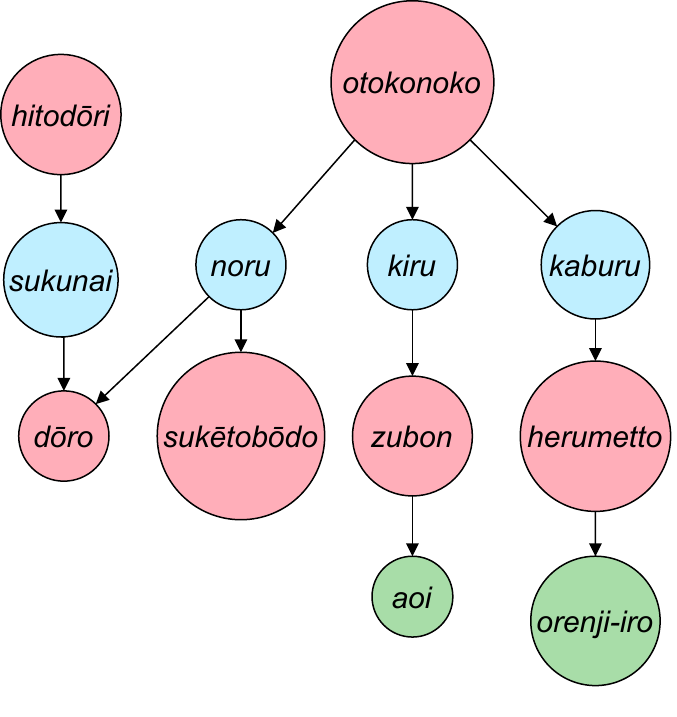}
    \subcaption{}
  \end{minipage}
  \caption{Example of an image and corresponding scene graph. The pink, green, and light blue nodes represent objects, attributes, and relationships, respectively, and the arrows represent dependencies. The caption is ``\textit{hitodōri no sukunaku natta dōro de, aoi zubon o kita otokonoko ga orenji-iro no herumetto o kaburi, sukētobōdo ni notte iru.}'' }
  \label{fig:sample}
  \vspace{-0.4cm}
\end{figure}

Fig. \ref{fig:sample} shows an example of an image and scene graph. Fig. \ref{fig:sample} (b) shows a scene graph obtained from the description ``\textit{hitodōri no sukunaku natta dōro de, aoi zubon o kita otokonoko ga orenji-iro no herumetto o kaburi, sukētobōdo ni notte iru.}'' (``on a deserted street, a boy in blue pants and an orange helmet rides a skateboard.'') for Fig. \ref{fig:sample} (a).
The pink, green, and light blue nodes represent objects, attributes, and relationships, respectively, and the arrows represent dependencies.

\vspace{-0.2cm}
\subsection{PAS-Based Scene Graph Parser (PAS-SGP)}

The input of PAS-SGP is generated caption $\hat{y}$ and the output is scene graph $G(\hat{y})$.
First, the morphological analyzer, syntactic analyzer, and predicate-argument structure analyzer\footnote{In this study, we employed the tools JUMAN++ \cite{jumanpp} and KNP \cite{kmp}.} extract the PAS and dependencies from $\hat{y}$.
Next, scene graph $\mathcal{G} \left\langle O(\hat{y}),E(\hat{y}),K(\hat{y}) \right\rangle $ is generated from the PAS and the dependencies by a rule-based method based on 10 case markers.
Note that the 10 case markers are: \textit{ga, wo, ni, to, de, kara, yori, he, made} and deep cases (e.g., temporal case) \cite{deep-case}.
Our parser directly extracts objects, relations, and attributes from the PAS and the dependencies. To parse them, we have defined a total of 13 dependency patterns. These patterns are designed to encapsulate the following constructions and phenomena:
\begin{itemize}
    \vspace{-4mm}
    \setlength{\parskip}{0cm}
    \setlength{\itemsep}{0cm}
    \item Subject–object–verb constructions
    \item Possessive constructions
    \item Prepositional phrases
    \item Clausal modifiers of nouns
    \item Adjectival modifiers
    \item Postpositional phrases
\end{itemize}

Furthermore, it is important to consider zero pronouns\cite{zero-pronouns} when comparing two sentences.
Consider two sentences A and B with the same meaning, but only sentence A contains a zero pronoun. Sentence A contains a relation that includes zero pronouns, which does not match any relation in sentence B. Hence, even though sentences A and B have the same meaning, not all relations match because of the zero pronoun.
Therefore, without careful handling, it is not possible to determine a suitable match.

To alleviate this issue, the proposed method performs heuristic zero anaphora resolution. Algorithm \ref{algo:zero-anaphora} shows the node completion algorithm of zero pronouns ($\phi$ represents a zero pronoun). 

\vspace{-0.1cm}
\begin{algorithm}[H]
    \caption{Node completion for zero pronouns}
    \label{algo:zero-anaphora}
    \begin{algorithmic}
        \Require $r \in R$
        \For {$o \in \mathrm{get\_objects(}r\mathrm{)}$}
            \If {$\mathrm{Rel} \left\langle \phi,r,o \right\rangle$ is found} \Comment{indegree is $0$}
                \For {$ \mathrm{Rel} \left\langle o',r',o \right\rangle \in \mathrm{find\_rel}(o)$}
                \State $\mathrm{Rel} \left\langle \phi,r,o \right\rangle \gets \mathrm{Rel} \left\langle o',r,o \right\rangle$
                \EndFor
            \EndIf
        \EndFor
    \end{algorithmic}
\end{algorithm}

\subsection{Graph Analyzer (GA)}
The inputs of GA are $\{G(y_{i,j})\}_{j=1}^N$ and $G(\hat{y})$, where $\{G(y_{i,j})\}_{j=1}^N$ is a set of scene graphs obtained from $\{y_{i,j}\}_{j=1}^N$.
First, GA expands $\{G(y_{i,j})\}_{j=1}^N$ and $G(\hat{y})$ by introducing synonym nodes as follows:
Suppose that objects $o_1$ and $o_2$ are connected by relation $r$. Given that $S(x)$ denotes the set of synonyms of $x$, our method generates new relations $\mathrm{Rel} \left\langle o'_1,r',o'_2 \right\rangle$, where $o'_1 \in S(o_1), o'_2 \in S(o_2),$ and $r' \in S(r)$. In other words, it adds new nodes $o \in S(o_1) \cup S(o_2)$ and $n$ new edges to the scene graph, where $n$ denotes $\left(|S(o_1)| + |S(o_2)|\right) \times |S(r)|$.
Note that we use the Japanese WordNet\cite{ja-wordnet} to obtain the set of synonyms.
We name this process \textit{graph extension}.

Next, GA merges scene graphs $\{G(y_{i,j})\}_{j=1}^N$ into a single graph.
Specifically, GA transforms $\mathcal{G}  \left\langle O(y_{i,j}),E(y_{i,j}),K(y_{i,j}) \right\rangle $ into:
\begin{align}
&G(\bm{y}_i) \; \delequal \nonumber \\ 
&\qquad \mathcal{G}  \left\langle \{O(y_{i,j})\}_{j=1}^N,\{E(y_{i,j})\}_{j=1}^N,\{K(y_{i,j})\}_{j=1}^N \right\rangle , \nonumber
\end{align}
where $\bm{y}_i$ denotes $\{y_{i,j}\}_{j=1}^N$.
To evaluate matching between both scene graphs in the range of $[0,1]$, GA computes the $F_1$ score from $G'(\hat{y})$ and $G(\bm{y}_i)$. The $F_1$ score is appropriate because it can take into account the difference in size between $G'(\hat{y})$ and $G(\bm{y}_i)$.
Precision $P$, recall $R$, and JaSPICE are defined as follows:

\vspace{-5mm}
\jot=0.3cm
\begin{gather}
P(\hat{y}, \bm{y_i}) = \frac{ \left|T\left(G'\left(\hat{y}\right) \right) \otimes T\left( G\left(\bm{y}_i\right)\right) \right|}{\left|T\left(G'\left(\hat{y}\right)\right)\right|} , \nonumber \\
R(\hat{y}, \bm{y_i}) = \frac{\left|T\left(G'\left(\hat{y}\right)\right) \otimes T\left(G\left(\bm{y}_i\right)\right)\right|}{ \left|T(G\left(\bm{y}_i)\right)\right|} , \nonumber \\ 
\mathrm{JaSPICE}(\hat{y}, \bm{y_i}) = \frac{2 \cdot P(\hat{y}, \bm{y_i}) \cdot R(\hat{y}, \bm{y_i})}{  P(\hat{y}, \bm{y_i}) + R(\hat{y}, \bm{y_i})} . \nonumber
\end{gather}

Note that we define $T(G(x))$ as:
\begin{align}
T(G(x)) \delequal O(x) \cup E(x) \cup K(x), \nonumber
\end{align}
and $\otimes$ denotes a function that returns matching tuples in two scene graphs.

\section{Experiments
\label{sec:experiments}
}

\subsection{Setup}

We conducted experiments to compare JaSPICE with existing automatic evaluation metrics.
In the experiments, we calculated the correlation coefficients between automatic evaluation metrics and human evaluation.
For the evaluation, we used outputs from the image captioning models, $\{y_i\}$ and $\{y_\mathrm{rand}\}$, obtained from STAIR Captions\cite{stair} and PFN-PIC\cite{pfn-pic}, which consisted of 21,227 and 1,920 captions, respectively.
Note that $\{y_i\}$ was randomly selected from $\{y_{i,j}\}_{j=1}^M$, and $y_\mathrm{rand}$ was randomly selected from all of $\{y_{i,j} | i=1,...,N,\; j=1,...,M \}$, where $M$ is the number of captions included per image.
We used a crowdsourcing service to collect human evaluations from 500 evaluators (The details are explained in Section \ref{new-dataset}).
For a given image, the human evaluators rated the appropriateness of its caption on a five-point scale.
To evaluate the proposed metric, we calculated the correlation coefficient (Pearson/Spearman/Kendall's correlation coefficient) between $\{s_J^{(i)}\}_{i=1}^N$ and $\{s_H^{(i)}\}_{i=1}^N$, where $s_J^{(i)}$ and $s_H^{(i)}$ denote the JaSPICE for the $i$-th caption and the human evaluation for the $i$-th caption, respectively.

Although there were problems with translation quality and speed, it was technically possible to compute SPICE by translating $\hat{y}$ and $\{y_{i,j} | i=1,...,N,\; j=1,...,M \}$ into English.
Thus, we conducted a comparison experiment between the proposed metric and SPICE obtained in this manner.
In the experiments, we calculated the correlation coefficient between the human evaluation and SPICE obtained from the English translation.
To avoid quality issues specific to a single machine translation, we performed the English translations using multiple approaches.
Specifically, we used a vanilla Transformer trained on JParaCrawl\cite{jparacrawl} and a proprietary machine translation system \footnote{We used \href{https://deepl.com}{DeepL} as a proprietary machine translation tool.}.

In this study, we used caption-level correlation $f(\{s_J^{(i)}\}_{i=1}^N, \{s_H^{(i)}\}_{i=1}^N)$ for the evaluation. In \cite{spice}, caption-level correlation $f(\{s_S^{(i)}\}_{i=1}^N, \{s_H^{(i)}\}_{i=1}^N)$ and system-level correlation $f(\{\bar{s}_S^{(j)}\}_{j=1}^{J}, \{\bar{s}_H^{(j)}\}_{j=1}^{J})$ were used to evaluate the automatic evaluation metric, where $f, s_S^{(i)}$, and $J$ denote the correlation coefficient function, SPICE for the $i$-th caption and the number of models, respectively.
However, because $J$ is generally very small, it is not appropriate to use system-level correlation $f(\{\bar{s}_S^{(j)}\}_{j=1}^{J}, \{\bar{s}_H^{(j)}\}_{j=1}^{J})$ for the evaluation. 
In fact, in \cite{kilickaya-etal-2017-evaluating}, the authors also used only the correlation coefficient per caption for the evaluation.

\begin{table}[t]
\centering
\caption{Correlation coefficients between each automatic evaluation metric and the human evaluation for STAIR Captions.}
\normalsize
\begin{tabular}{lccc}
\hline
{Metric} & {Pearson} & {Spearman} & {Kendall} \\ \hline
{BLEU} & {0.296} & {0.343} & {0.260} \\
{ROUGE} & {0.366} & {0.340} & {0.258} \\
{METEOR} & {0.345} & {0.366} & {0.279} \\
{CIDEr} & {0.312} & {0.355} & {0.269} \\
{JaSPICE} & {\textbf{0.501}} & \textbf{0.529} & \textbf{0.413} \\ \hline
{$r_\mathrm{human}$} & {0.759} & {0.750} & {0.669} \\ \hline
\end{tabular}
\label{tab:corr-results}
\end{table}

\begin{table}[t]
\centering
\caption{Comparison between JaSPICE and SPICE in terms of correlation with human evaluation for STAIR Captions.}
\normalsize
\begin{tabular}{lccc}
\hline
{Metric} & {Pearson} & {Spearman} & {Kendall} \\ \hline
{$\mathrm{SPICE_{service}}$} & {0.488} & {0.515} & {0.402} \\ 
{$\mathrm{SPICE_{trm}}$} & {0.491} & {0.516} & {0.403} \\
{JaSPICE} & {\textbf{0.501}} & \textbf{0.529} & \textbf{0.413} \\ \hline
\end{tabular}
\label{tab:corr-results2}
\vspace{-0.5cm}
\end{table}

\vspace{-0.25cm}
\subsection{Corpora and Models}

In this study, we used STAIR Captions and PFN-PIC as corpora.
STAIR Captions is a large-scale Japanese image-caption corpus, and PFN-PIC is a corpus for a robotic system, which contains object manipulation instructions in English and Japanese.
We adopted these corpora because STAIR Captions is a standard Japanese image caption corpus based on MS-COCO images, and PFN-PIC is a standard dataset that comprises images and a set of instructions in Japanese for a robotic system.


To evaluate the proposed metric on STAIR Captions, we used a set of 10 standard models, including SAT\cite{sat}, ORT\cite{ort}, $\mathcal{M}^2$-Transformer\cite{m2trm}, DLCT\cite{dlct},  ER-SAN\cite{er-san}, $\mathrm{ClipCap_{mlp}}$\cite{clipcap}, $\mathrm{ClipCap_{trm}}$, and $\mathrm{Transformer}_{L = 3,6,12}$\cite{transformer}. We trained these models on STAIR Captions from scratch.
Additionally, to evaluate the proposed metric on PFN-PIC, we used a set of 3 standard models, including CRT\cite{crt}, ORT, and SAT.
The details are explained in Appendix \ref{appendix:setup}.

\vspace{-0.25cm}
\subsection{Experimental Results: STAIR Captions}

To validate the proposed metric, we experimentally compared it with the baseline metrics using their correlation with human evaluation.


Table \ref{tab:corr-results} shows the quantitative results for the proposed metric and baseline metrics on STAIR Captions.
Note that $r_\mathrm{human}$ is explained in Section \ref{new-dataset}.
For the baseline metrics, we used BLEU\cite{bleu}, ROUGE\cite{rouge}, METEOR\cite{meteor} and CIDEr\cite{cider}, which are standard automatic evaluation metrics for image captioning.

Table \ref{tab:corr-results} shows that the Pearson, Spearman, and Kendall correlation coefficients between JaSPICE and the human evaluation were $0.501, 0.529$ and $0.413$, respectively, which indicates that JaSPICE outperformed all the baseline metrics.

Table \ref{tab:corr-results2} shows a comparison between JaSPICE and SPICE in terms of correlation with human evaluation.
Note that $\mathrm{SPICE_{trm}}$ and $\mathrm{SPICE_{service}}$ denote SPICE calculated from English translations by Transformer trained on JParaCrawl and a proprietary machine translation system, respectively.
Table \ref{tab:corr-results2} indicates that the Pearson, Spearman, and Kendall correlation coefficients between JaSPICE and the human evaluation were $0.501, 0.529$ and $0.413$, respectively. Thus, JaSPICE outperformed $\mathrm{SPICE_{trm}}$ by $0.010, 0.013$, and $0.010$ points for each correlation coefficient, respectively. Similarly, JaSPICE outperformed $\mathrm{SPICE_{service}}$ by $0.013, 0.014$, and $0.011$ points.

\begin{figure*}[t]
  \begin{minipage}[b]{0.24\linewidth}
    \centering
    \includegraphics[height=6cm,trim={0 0 {.24\linewidth} 0},clip]{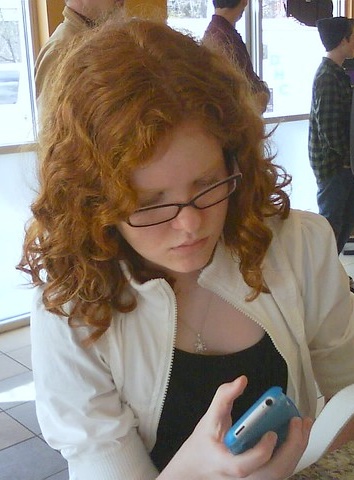}
  \end{minipage}
  \begin{minipage}[b]{0.24\linewidth}
    \centering
    \includegraphics[height=6cm]{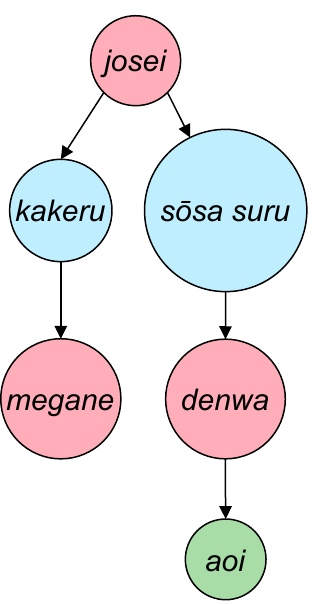}
  \end{minipage}
  \begin{minipage}[b]{0.24\linewidth}
    \centering
    \includegraphics[height=6cm,trim={0 0 {.24\linewidth} 0},clip]{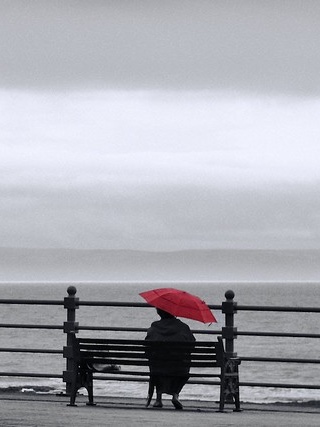}
  \end{minipage}
  \begin{minipage}[b]{0.24\linewidth}
    \centering
    \includegraphics[height=6cm]{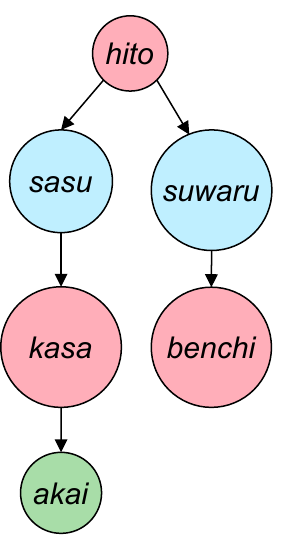}
  \end{minipage}
    \\ 
  \begin{minipage}[b]{0.5\linewidth}
    \centering
    (a)
  \end{minipage}
    \begin{minipage}[b]{0.5\linewidth}
    \centering
    (b)
  \end{minipage}
  \caption{Image and scene graph for successful cases for STAIR Captions. (a) $\hat{y}_i$: ``\textit{megane o kaketa josei ga aoi denwa o sōsa shite iru}'' (``a woman wearing glasses is operating a blue cell phone''), $s_H^{(i)} = 5, \mathrm{JaSPICE}(\hat{y},\bm{y}_i)  = 0.526 > \tau_S$; and (b) $\hat{y}_j$: ``\textit{akai kasa o sashita hito ga benchi ni suwatte iru}'' (``a person with a red umbrella is sitting on a bench''), $s_H^{(j)} = 5, \mathrm{JaSPICE}(\hat{y},\bm{y}_j)  = 0.632 > \tau_S$.
  }
  \label{fig:success-sample}
  \vspace{-0.5cm}
\end{figure*}

Fig. \ref{fig:success-sample} show successful examples of the proposed metric for STAIR Captions.
Fig. \ref{fig:success-sample} (a) illustrates an input image and its corresponding scene graph for $\hat{y}_k$ ``\textit{megane o kaketa josei ga aoi denwa o sōsa shite iru}'' (``a woman wearing glasses is operating a blue cell phone'').
For this sample, $y_{i,1}$ was ``\textit{josei ga aoi sumātofon o katate ni motte iru}'' (``woman holding blue smartphone in one hand'').
Regarding this sample, $\mathrm{JaSPICE}(\hat{y},\bm{y}_i)$ and $s_H^{(i)}$ were $0.588$ and $5$, respectively. 
In the STAIR Captions test set, 33.6\% of the total samples were rated as $s_H^{(i)} = 5$, whereas the top 33.6\% score in $\{\mathrm{JaSPICE}(\hat{y},\bm{y}_k)\}_{k=1}^N$ was observed to be $\tau_S = 0.207$. This sample satisfies $\mathrm{JaSPICE}(\hat{y},\bm{y}_i) > \tau_S$, suggesting that our metric generated an appropriate score for this sample.

Similarly, Fig. \ref{fig:success-sample} (b) shows an input image and scene graph for $\hat{y}_j$ ``\textit{akai kasa o sashita hito ga benchi ni suwatte iru}'' (``a person with a red umbrella is sitting on a bench'').
For Fig. \ref{fig:success-sample} (b), $y_{j,1}$ was ``\textit{akai kasa o sashite suwatte umi o mite iru}'' (``sitting with a red umbrella, looking out to sea.''), and regarding this sample, $\mathrm{JaSPICE}(\hat{y},\bm{y}_j)$ and $s_H^{(j)}$ were $0.632 
 (>\tau_S)$ and $5$, respectively. 
These results indicate that the proposed metric generated appropriate scores for STAIR Captions.


\begin{figure}[t]
  \begin{minipage}[b]{\linewidth}
    \centering
    \includegraphics[width=\linewidth]{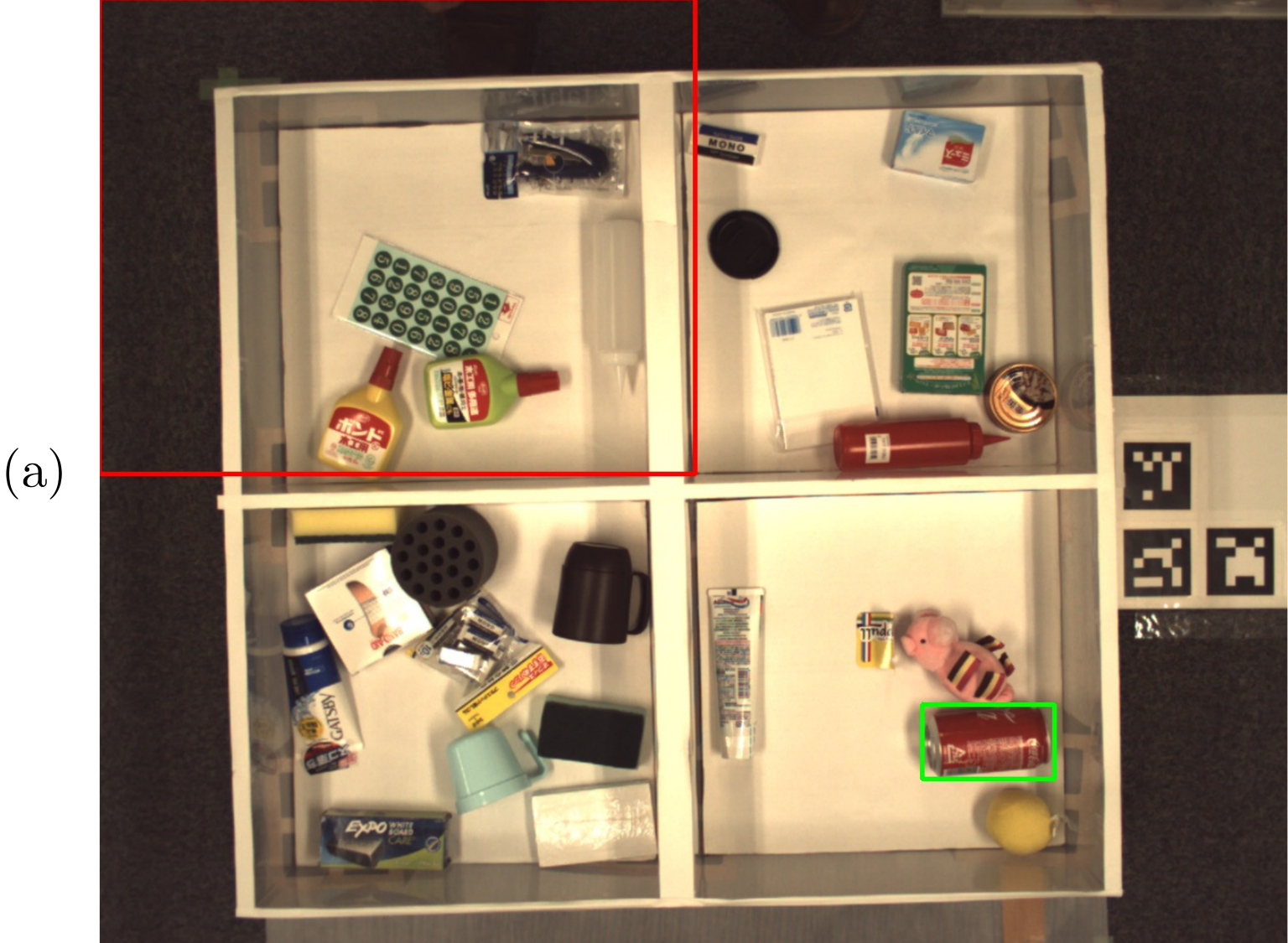}
  \end{minipage}
    \begin{minipage}[b]{\linewidth}
    \centering
    \includegraphics[width=\linewidth]{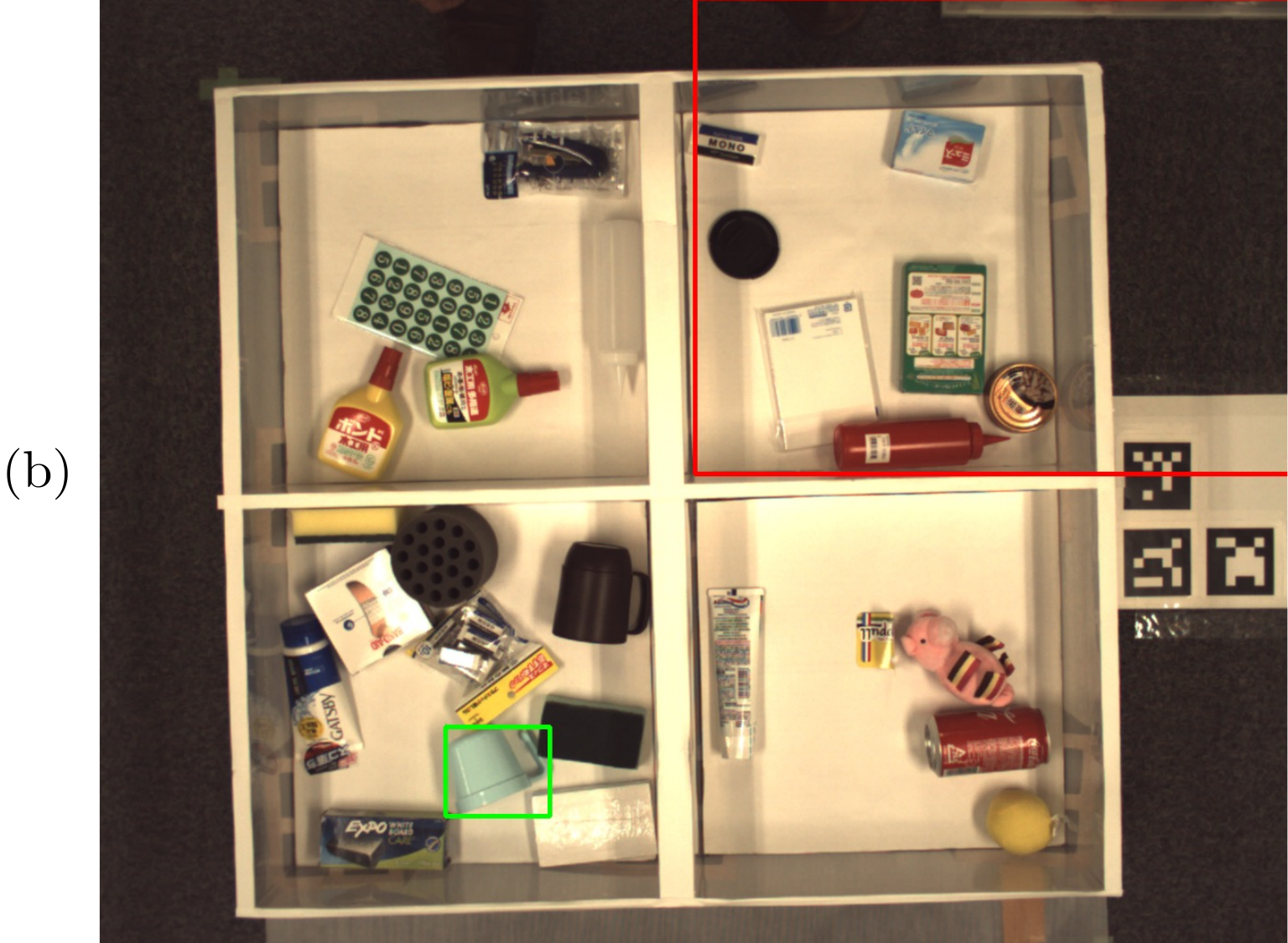}
  \end{minipage}
  \caption{Image for successful cases for PFN-PIC. (The green and red boxes in the figure
represent the target object and destination, respectively.) (a) $\hat{y}_i$: ``move the can of Coke in the box in the bottom right to the box in the top left'', $s_H^{(i)} = 5, \mathrm{JaSPICE}(\hat{y},\bm{y}_i) = 0.870 > \tau_P$; and (b) $\hat{y}_j$ ``move the blue cup to the box in the top right-hand corner'', $s_H^{(j)} = 5, \mathrm{JaSPICE}(\hat{y},\bm{y}_j)  = 0.385 > \tau_P$. Scene graphs for these samples are shown in the Appendix \ref{appendix:pfnpic-results}.
  }
  \label{fig:success-sample-pfnpic}
  \vspace{-0.5cm}
\end{figure}


\vspace{-0.25cm}
\subsection{Experimental Results: PFN-PIC}

Table \ref{tab:corr-results-pfnpic} shows the quantitative results for the proposed and baseline metrics for PFN-PIC.
Table \ref{tab:corr-results-pfnpic} indicates that the Pearson, Spearman, and Kendall correlation coefficients between JaSPICE and the human evaluation were $0.572$, $0.587$, and $0.452$, respectively, which indicates that JaSPICE outperformed all the baseline metrics.

Table \ref{tab:corr-results2-pfnpic} shows the correlation coefficients between JaSPICE and the human evaluation for the PFN-PIC dataset. The results indicate that JaSPICE also outperformed both $\mathrm{SPICE_{trm}}$ and $\mathrm{SPICE_{service}}$ on PFN-PIC.

\begin{table}[t]
\centering
\caption{Correlation coefficients between each evaluation metric and human evaluation for PFN-PIC.}
\normalsize
\begin{tabular}{lccc}
\hline
{Metric} & {Pearson} & {Spearman} & {Kendall} \\ \hline
{BLEU} & {0.484} & {0.466} & {0.352} \\
{ROUGE} & {0.500} & {0.474} & {0.365} \\
{METEOR} & {0.423} & {0.457} & {0.352} \\
{CIDEr} & {0.416} & {0.462} & {0.353} \\
{JaSPICE} & \textbf{0.572} & \textbf{0.587} & \textbf{0.452} \\ \hline
\end{tabular}
\label{tab:corr-results-pfnpic}
\end{table}

\begin{table}[t]
\centering
\caption{Comparison of JaSPICE and SPICE in terms of correlation with human evaluation for PFN-PIC.}
\normalsize
\begin{tabular}{lccc}
\hline
{Metric} & {Pearson} & {Spearman} & {Kendall} \\ \hline
{$\mathrm{SPICE_{service}}$} & {0.416} & {0.418} & {0.316} \\ 
{$\mathrm{SPICE_{trm}}$} & {0.427} & {0.420} & {0.317} \\
{JaSPICE} & \textbf{0.572} & \textbf{0.587} & \textbf{0.452} \\ \hline
\end{tabular}
\label{tab:corr-results2-pfnpic}
\end{table}

\begin{table}[t]
\centering
\setlength{\tabcolsep}{2.5pt}
\caption{Results of the ablation study (P: Pearson, S: Spearman, K: Kendall, $M$: the number of samples for which $\mathrm{JaSPICE}(\hat{y},\bm{y}_i) = 0$).}
\normalsize
\begin{tabular}{ccccccc}
\hline
{\small Metric} & {Parser} & {\small \begin{tabular}{c}Graph\\Extension\end{tabular}} & {P} & {S} &
{K} & {$M$} \\ \hline
{(i)} & {UD} & {} & {0.398} & {0.390} & {0.309} & {1465} \\
{(ii)} & {UD} & {$\checkmark$} & {0.399} & {0.390} & {0.309} & {1430} \\
{(iii)} & {JaSGP} & {} & {0.493} & {0.524} & {0.410} & {1417} \\
{(iv)} & {JaSGP} & {$\checkmark$} & {\textbf{0.501}} & \textbf{0.529} & \textbf{0.413} & \textbf{1346} \\ \hline
\end{tabular}
\label{tab:ablation}
\vspace{-0.6cm}
\end{table}

Fig. \ref{fig:success-sample-pfnpic} shows successful examples of the proposed metric for PFN-PIC.
Note that the green and red boxes in the figure represent the target object and destination, respectively.
Fig. \ref{fig:success-sample-pfnpic} (a) illustrates an input image and its corresponding scene graph for $\hat{y}_i$ ``\textit{migishita no hako no naka no kōra no kan o, hidariue no hako ni ugokashite kudasai}'' (``move the can of Coke in the box in the bottom right to the box in the top left'').
Regarding Fig. \ref{fig:success-sample-pfnpic} (a), $y_{i,1}$ was ``\textit{kōra no kan o, hidariue no kēsu ni ugokashite chōdai}'' (``move the can of Coke to the case in the top left-hand corner'').
For this sample, $\mathrm{JaSPICE}(\hat{y},\bm{y}_i)$ and $s_H^{(i)}$ were $0.870$ and $5$, respectively.
In the PFN-PIC test set, 41.2\% of the total samples were rated as $s_H^{(i)} = 5$, whereas the top 41.2\% score in $\{\mathrm{JaSPICE}(\hat{y},\bm{y}_k)\}_{k=1}^N$ was observed to be $\tau_P = 0.276$. This sample satisfies $\mathrm{JaSPICE}(\hat{y},\bm{y}_i) > \tau_P$, suggesting that our metric generated an appropriate score for this sample.


Similarly, Fig. \ref{fig:success-sample-pfnpic} (b) shows an input image and a scene graph for $\hat{y}_j$ ``\textit{mizuiro no kappu o, migiue no hako ni ugokashite kudasai}'' (``move the blue cup to the box in the top right-hand corner.'').
Regarding Fig. \ref{fig:success-sample-pfnpic} (b), $y_{j,1}$ was ``\textit{hidarishita no hako no naka ni aru mizuiro no kappu o, migiue no hako ni ugokashite kudasai}'' (``move the blue cup from the bottom left box to the top right box''). For this sample, $\mathrm{JaSPICE}(\hat{y},\bm{y}_j)$ and $s_H^{(j)}$ were $0.385(>\tau_P)$ and $5$, respectively.
These results indicate that the proposed metric also generated appropriate scores for PFN-PIC.



\vspace{-0.25cm}
\subsection{Experimental Results: \textit{Shichimi}}
\label{new-dataset}

Although the above experiment was compared to baseline metrics, it is also important to compare metrics with $r_\mathrm{human}$, the correlation coefficient within human evaluations.
Hence, to calculate $r_\mathrm{human}$, we constructed the \textbf{\textit{Shichimi}} (Subject Human evaluatIons of CompreHensive Image captioning Model's Inferences) dataset containing a total of 103,170 human evaluations collected from $500$ evaluators.
The \textit{Shichimi} dataset, which includes images, captions, and human evaluations on a five-point scale, is a versatile resource that can be efficiently utilized to develop regression-based metrics such as COMET \cite{comet}.

We found $r_\mathrm{human}$ to be $0.759$ on the \textit{Shichimi} dataset.
The reason for $r_\mathrm{human}$ being less than $1.0$ is the variability among human evaluations within the same sample.
Here,  we define $r_\mathrm{human}$ as  $\mathbb{E}[\mathcal{R}(Y_i, Y_j)]$, where $Y_i$ and $\mathcal{R}$ denote the human evaluation vector by the $i$-th user and the correlation coefficient function, respectively.
$r_\mathrm{human}$ is considered to be a virtual upper bound on the performance of the automatic evaluation metrics.
Among the baseline metrics, the correlation coefficient of ROUGE, which performed best, was $0.366$. This was a difference of $0.393$ from $r_\mathrm{human}$, indicating that the use of baseline metrics for the evaluation of image captioning could be problematic. 
Meanwhile, the difference between the correlation coefficient in JaSPICE and $r_\mathrm{human}$ was $0.258$. Although this shows an improvement over the baseline metrics, there remains scope for further enhancement (Error analysis and discussion can be found in Appendix \ref{appendix:error-analysis}).

\vspace{-0.25cm}
\subsection{Ablation Studies}


We defined two conditions for ablation studies. 
Table \ref{tab:ablation} shows the results of the ablation study.
For each condition, we examined not only the correlation coefficient but also the number of samples $M$ for which $\mathrm{JaSPICE}(\hat{y},\bm{y}_i) = 0$. This is because JaSPICE might produce a zero output when no matched pairs are found during the comparison between pairs in $T(G'(\hat{y}))$ and $T(G(\bm{y}_i))$.

\paragraph{Scene Graph Parser Ablation}

We replaced PAS-SGP with a scene graph parser based on UD (UD parser) to investigate the performance of PAS-SGP.
In comparison with Metric (iv), under Metric (ii), the values of the Pearson, Spearman, and Kendall correlation coefficients were $0.102, 0.139$, and $0.104$ points lower, respectively.
Furthermore, there were $119$ fewer samples for $M$.
This indicates that the introduction of the PAS-SGP contributed the most to performance. 

\paragraph{Graph Extension Ablation}
We investigated the influence on performance when the graph extension was removed.
A comparison between Metric (i) and (iv), in addition to (iii) and (iv), suggests that the introduction of graph extensions also contributed to the performance improvement.


\section{Conclusions}
In this study, we proposed JaSPICE, which is an automatic evaluation metric for image captioning models in Japanese.
The following contributions of this study can be emphasized:
\begin{itemize}
    \vspace{-1mm}
    \setlength{\parskip}{0cm}
    \setlength{\itemsep}{0cm}
    \item We proposed JaSPICE, which is an automatic evaluation metric for image captioning models in Japanese.
    \item Unlike SPICE, we proposed a rule-based scene graph parser PAS-SGP using dependencies and PAS.
    \item We introduced graph extension using synonyms to take synonyms into account in the evaluation.
    \item We constructed the \textit{Shichimi} dataset, which contains a total of 103,170 human evaluations collected from 500 evaluators.
    \item Our method outperformed SPICE calculated from English translations and the baseline metrics on the correlation coefficient with the human evaluation.
\end{itemize}

In future studies, we will extend our method by taking into account hypernyms and hyponyms.


\section*{ACKNOWLEDGMENT}
This work was partially supported by JSPS KAKENHI Grant Number 23H03478, JST CREST, and NEDO.
\bibliographystyle{acl_natbib}
\bibliography{reference}

\clearpage
\appendix

\section{Corpora and Systems}
\label{appendix:setup}
The STAIR Captions\cite{stair} contains 5 captions for each of 164,062 images, for a total of 820,310 captions. The vocabulary size is 35,642 and the average sentence length is 23.79. The captions were annotated by 2,100 Japanese speakers.

The PFN-PIC\cite{pfn-pic} is annotated by at least three annotators for each object and divided into training and validation sets. The training set consists of 1,180 images, 25,900 target objects, and 91,590 instructions, and the validation set consists of 20 images, 352 target objects, and 898 instructions.

In the experiments, we divided both STAIR Captions and PFN-PIC into training, validation, and test sets.
Note that STAIR Captions included $413{,}915$; $37{,}269$; and $35{,}594$ captions, and PFN-PIC included $81{,}087$; $8{,}774$; and $898$ samples, respectively.

To evaluate the proposed metric on STAIR Captions, we used a set of 10 standard models. Table \ref{tab:models} shows the systems used in the experiments.
Note that $\mathrm{ClipCap_{mlp}}$ and $\mathrm{ClipCap_{trm}}$ are variations of ClipCap that incorporate MLP and Transformer as Mapping Networks, respectively, whereas $\mathrm{Transformer}_L$ denotes $L$-layer Transformer models with Bottom-up features\cite{bottom-up} as inputs.

\begin{table}[H]
\centering
\caption{The system used in the experiments.}
\begin{tabular}{cc}
\hline
{System} & {Citation} \\ \hline
SAT & \cite{sat} \\
ORT&\cite{ort} \\
{$\mathrm{Transformer}_{L = 3}$}&\cite{transformer}\\
{$\mathrm{Transformer}_{L = 6}$}&\cite{transformer}\\
{$\mathrm{Transformer}_{L = 12}$}&\cite{transformer}\\
$\mathcal{M}^2$-Transformer&\cite{m2trm}\\
DLCT&\cite{dlct}\\
ER-SAN&\cite{er-san}\\
{$\mathrm{ClipCap_{mlp}}$}&\cite{clipcap}\\
{$\mathrm{ClipCap_{trm}}$}&\cite{clipcap}\\
CRT&\cite{crt}\\
Human&{---} \\
Random&{---} \\
\hline
\end{tabular}
\label{tab:models}
\end{table}


\section{Applications of image captioning}
Numerous studies have been conducted in the field of image captioning\cite{sat,ort,m2trm,dlct,er-san}, a crucial area of research that has been further extended and applied in the sphere of robotics\cite{mabn,alleviating,crt}.
Multi-ABN\cite{mabn} is a model for generating fetching instructions for domestic service robots using multiple images from various viewpoints. ABEN\cite{alleviating} is a model that extends Multi-ABN and introduces linguistic and generative branches to model relationships between subwords, thus achieving subword-level attention. CRT\cite{crt} is a model for generating fetching instructions including the spatial referring expressions of target objects and destinations. It introduces Transformer-based encoder-decoder architecture to fuse the visual and geometric features of the objects in images.

\vspace{-2.5mm}
\section{Experimental Results: PFN-PIC}
\label{appendix:pfnpic-results}
\vspace{-2.5mm}

\begin{figure}[H]
  \begin{minipage}[b]{\linewidth}
    \centering
    \includegraphics[height=5cm]{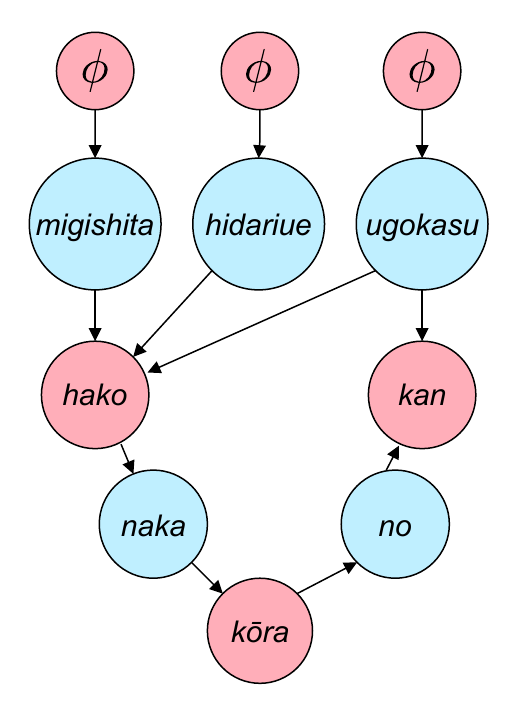}
  \end{minipage}
  \begin{minipage}[b]{\linewidth}
    \centering
    \includegraphics[height=5cm]{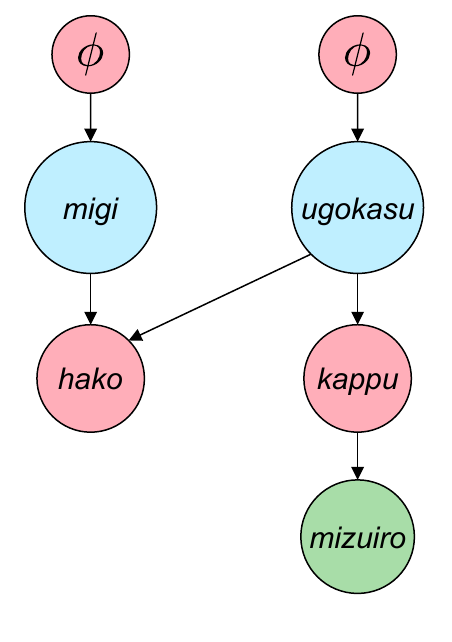}
  \end{minipage}
  \caption{Scene graph for successful cases for PFN-PIC. $\phi$ represents a zero pronoun.  (a) $\hat{y}_i$: ``\textit{migishita no hako no naka no kōra no kan o, hidariue no hako ni ugokashite kudasai}'' (``move the can of Coke in the box in the bottom right to the box in the top left''), $s_H^{(i)} = 5, \mathrm{JaSPICE}(\hat{y},\bm{y}_i) = 0.870 > \tau_P$; and (b) $\hat{y}_j$ ``\textit{mizuiro no kappu o, migiue no hako ni ugokashite kudasai}'' (``move the blue cup to the box in the top right-hand corner''), $s_H^{(j)} = 5, \mathrm{JaSPICE}(\hat{y},\bm{y}_j)  = 0.385 > \tau_P$. 
  }
  \label{fig:scenegraphs-pfnpic}
\end{figure}

Fig \ref{fig:scenegraphs-pfnpic} shows the scene graphs for the samples in Fig \ref{fig:success-sample-pfnpic}.

\section{Failure Cases}
\begin{figure}[H]
  \begin{minipage}[b]{0.75\linewidth}
    \centering
    \includegraphics[width=\linewidth]{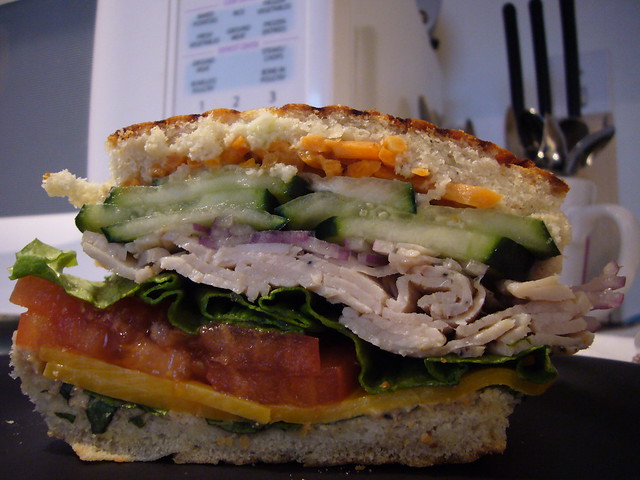}
  \end{minipage}
  \begin{minipage}[b]{0.165\linewidth}
    \centering
    \includegraphics[width=\linewidth]{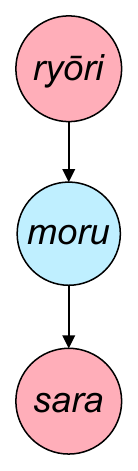}
  \end{minipage}
  \caption{Image and scene graph in failed cases for STAIR Captions; $\hat{y}_k$ : ``\textit{sara ni ryōri ga mora rete iru}'' (``food is served on a plate''), $s_H^{(k)} = 5, \mathrm{JaSPICE}(\hat{y},\bm{y}_k)  = 0 < \tau_S$. 
  }
  \label{fig:failure-sample}
  \vspace{-4mm}
\end{figure}

Fig. \ref{fig:failure-sample} shows an unsuccessful example of the proposed metric.
Fig. \ref{fig:failure-sample} illustrates an input image and its corresponding scene graph for $\hat{y}_k$ ``\textit{sara ni ryōri ga mora rete iru}'' (``food is served on a plate'').
For Fig. \ref{fig:failure-sample}, $y_{k,1}$ was ``\textit{pan ni hamu to kyūri to tomato to chīzu ga hasamatte iru}'' (``bread with ham, cucumber, tomato and cheese.'').
For this sample, JaSPICE was $0$ even though $s_H^{(k)}$ was $5$.
In this case, $y_{k,1}$ used the terms ``bread'' and ``ham'' whereas $\hat{y}$ used the hypernym ``food'', which resulted in a lower output score because of the mismatch in wording.

\section{Error Analysis and Discussion}
\label{appendix:error-analysis}


We define the failed cases of the proposed metric as a sample that satisfies
$ \left|\frac{s_H^{(i)}}{\underset{i}{\max}\; s_H^{(i)}} - \frac{s_J^{(i)}}{\underset{i}{\max}\; s_J^{(i)}}\right| \geq \theta . \nonumber $
In this study, we set $\theta = 1$ and there were $130$ failed samples in the test set.

We investigated $100$ out of $130$ failed samples. Table \ref{tab:failure} categorizes the failure cases. The causes of failure can be divided into five groups:

\begin{enumerate}
\vspace{-2mm}
\setlength{\parskip}{0cm}
\setlength{\itemsep}{0cm}

\item[(i) ] Word granularity differences in $\hat{y}$ and $\bm{y}_i$ : This refers to cases in which $\bm{y}_i$ used a hyponym for a certain object, relation or attribute in the image, whereas $\hat{y}$ used a hypernym. 
In the example shown in Fig. \ref{fig:failure-sample}, the hyponym ``bread'' was represented by the hypernym ``food'' in $\hat{y}$.

\item[(ii) ] Difference in focus: This refers to the case in which the focuses of $\bm{y}_i$ and $\hat{y}$ were different. 
Both captions were appropriate but focused on different aspects, leading to an inappropriate JaSPICE score.

\item[(iii) ] Comparison of sentences containing partially matching morphemes: For example, if $\hat{y}$ was a sentence containing ``tennis racket'' and $y_{i,1}$ was a sentence containing ``tennis,'' then scene graphs had fewer matching pairs, which resulted in an inappropriate JaSPICE.

\item[(iv) ] Erroneous evaluation: 
This refers to cases in which there was a discrepancy between $S_H^{(i)}$ and the quality of $\hat{y}_i$.

\item[(v)] Others: This category includes other errors.

\end{enumerate}

Table \ref{tab:failure} highlights the main bottleneck of the proposed method: the discrepancy in word granularity between $\hat{y}$ and $\bm{y}_i$. Therefore, we consider that the bottleneck can be reduced by the introduction of a model that takes into account the relation between hypernyms and hyponyms.

\begin{table}[H]
    \centering
    \caption{Categorization of failed samples.}
    \normalsize
    \begin{tabular}{cc}
    \hline
    {Error} & {\#Samples} \\ \hline
    {(i)} & {46} \\
    {(ii)} & {20} \\
    {(iii)} & {18} \\
    {(iv)} & {10} \\
    {(v) Others} & {6} \\ \hline
    \end{tabular}
    \label{tab:failure}
\end{table}

\section{Details of the \textit{Shichimi} Dataset}
\label{appendix:shichimi}

We removed inappropriate users from the \textit{Shichimi} dataset (e.g. users with extremely short response times \cite{spi} or those who only responded with the same values).

Table \ref{tab:shichimi-dist} shows the distribution of human evaluations on the \textit{Shichimi} dataset.

\begin{table}[H]
    \centering
    \caption{The distribution on the \textit{Shichimi} dataset.}
    \normalsize
    \begin{tabular}{lc}
    \hline
    {Score} & {\#Samples} \\ \hline
    {5 (Excellent)} & {31,809} \\
    {4 (Good)} & {21,857} \\
    {3 (Fair)} & {22,513} \\
    {2 (Poor)} & {12,873} \\
    {1 (Bad)} & {14,118} \\ \hline
    \end{tabular}
    \label{tab:shichimi-dist}
\end{table}

\end{document}